\documentclass{fairmeta}
\usepackage{caption}
\usepackage{amsmath}
\usepackage{enumerate} 
\usepackage{bm}
\usepackage{algorithm}
\usepackage{floatflt}
\usepackage{algpseudocode}
\usepackage{amsfonts}
\usepackage{amsthm}
\usepackage{newtxtt}
\usepackage{algorithm}
\usepackage{colortbl} 
\usepackage{cleveref}
\usepackage{diagbox} 
\usepackage[utf8]{inputenc}
\usepackage{textgreek}
\usepackage{colortbl}
\usepackage{nicematrix}
\usepackage{makecell}
\usepackage{float}
\usepackage{arydshln}
\usepackage[frozencache,cachedir=.]{minted}
\usepackage{caption}
\usepackage{subcaption}
\usepackage{tcolorbox}
\usepackage{amssymb}
\usepackage{xspace}
\usepackage{wrapfig}
\usepackage{adjustbox}
\usepackage{tabularx}
\usepackage{booktabs}
\usepackage{mathtools}
\usepackage{wrapfig}
\usepackage{amssymb}
\usepackage{graphicx}

\usepackage{silence}
\makeatletter
\patchcmd{\wrong@fontshape}{\@gobbletwo}{}{}{}
\makeatother
\definecolor{upColor}{RGB}{17,138,21}
\definecolor{downColor}{RGB}{174,36,67}

\newtheorem{theorem}{Theorem}[]

\newtheorem{remark1}[theorem]{Remark}

\title{Ruyi2.5 Technical Report}
\author[]{Huan Song}
\author[]{Shuyu Tian}
\author[]{Qingfei Zhao}
\author[]{Wenhao Hong}
\author[]{Jiang Liu}
\author[]{Ting Long}
\author[]{Jiawei Shao}
\author[]{Xuelong Li}
\affiliation[]{Institute of Artificial Intelligence (TeleAI), China Telecom}

\date{
   March 17, 2026 
}

\metadata[Correspondence to]{Xuelong Li (\email{xuelong\_li@ieee.org})}

\begin{document}

\abstract{
We present Ruyi2.5, a multimodal familial model built on the AI Flow framework. Extending Ruyi2's "Train Once, Deploy Many" paradigm to the multimodal domain, Ruyi2.5 constructs a shared-backbone architecture that co-trains models of varying scales within a single unified pipeline, ensuring semantic consistency across all deployment tiers. Built upon Ruyi2.5, Ruyi2.5-Camera model is developed as a privacy-preserving camera service system, which instantiates Ruyi2.5-Camera into a two-stage recognition pipeline: an edge model applies information-bottleneck-guided irreversible feature mapping to de-identify raw frames at the source, while a cloud model performs deep behavior reasoning—covering occlusion, multi-person interaction, and temporal dynamics over the anonymized representations. To accelerate reinforcement learning fine-tuning, we further propose Binary Prefix Policy Optimization (BPPO), which reduces sample redundancy via binary response selection and focuses gradient updates on response prefixes, achieving a 2–3× training speedup over GRPO. Experiments show Ruyi2.5 matches Qwen3-VL on the general multimodal benchmarks, while Ruyi2.5-Camera substantially outperforms Qwen3-VL on privacy-constrained surveillance tasks.
}

\maketitle

\section{Introduction}

Intelligent video analytics has become an indispensable component of public safety, healthcare monitoring, and smart-building management \citep{chen2024internvl,wang2024qwen2vl}. The ability to automatically detect risk behaviors—such as falls, physical altercations, or smoking in prohibited areas—can substantially reduce human oversight costs and emergency response latency. However, deploying surveillance systems in privacy-sensitive locations such as restrooms, locker rooms, and medical facilities raises profound ethical and legal concerns. Transmitting raw video streams to centralized cloud servers exposes personally identifiable information (PII), even when the downstream analysis is entirely benign in intent \citep{rigaki2023surveyprivacyattacksml}. This creates a fundamental tension that existing approaches have yet to fully resolve: privacy protection and semantic understanding appear to be competing objectives.

Existing privacy-preserving strategies broadly fall into three categories. Physical obfuscation methods rely on non-visual sensing modalities such as depth or infrared cameras to avoid capturing identifying visual details \citep{nakashima2024privacypreservingactionsurvey}. Cryptographic approaches encrypt data prior to transmission, incurring prohibitive computation overhead for real-time applications \citep{dowlin2016cryptonets}. Visual anonymization methods—including face blurring, body silhouetting, and skeleton extraction—remove identity-sensitive regions from video frames before analysis \citep{hukkelas2019deepprivacy,zhang2024pca}. While these strategies offer varying degrees of privacy protection, they share a fundamental limitation: the anonymization pipeline inevitably discards semantically meaningful content alongside identity cues, degrading downstream recognition accuracy. Moreover, most methods that transmit intermediate feature representations to the cloud remain vulnerable to model inversion and reconstruction attacks \citep{fredrikson2015modelinversion}, undermining the privacy guarantee they claim to provide.
Moreover, a largely orthogonal challenge arises from the deployment scale of modern multimodal large language models (MLLMs). Models such as LLaVA \citep{liu2023visualinstructiontuning}, InternVL \citep{chen2024internvl}, and the Qwen-VL family \citep{bai2023qwenvl,wang2024qwen2vl} have demonstrated remarkable cross-modal understanding and reasoning capabilities. Moreover, this tier-specific training paradigm incurs substantial computational overhead, as each model size requires an independent, resource-intensive training process from scratch. Neither approach exploits the structural relationship between models of different scales: that a compact model trained on the same data distribution as its larger counterpart should share, rather than relearn, the underlying semantic space.

The training pipeline for such models introduces a further difficulty. Reinforcement learning from verifiable rewards (RLVR), as operationalized by GRPO \citep{shao2024deepseekmath} in DeepSeekMath and DeepSeek-R1 \citep{deepseekai2025deepseekr1}, has emerged as a powerful framework for instilling complex reasoning into LLMs. However, the standard group sampling strategy generates G responses per prompt and optimizes over all of them, producing correlated gradient signals that inflate effective batch size without proportionally increasing informational diversity. Simultaneously, backpropagating through the full token sequence of long reasoning chains incurs prohibitive memory and wall-clock costs \citep{yu2025dapo}, forming a practical bottleneck for multimodal RLVR at scale.

Building upon the AI Flow theoretical framework \citep{shao2025ai}, we introduce Ruyi2.5, a multimodal familial model that addresses all three challenges within a unified architecture. Ruyi2 \citep{song2026ruyi2} established the "Train Once, Deploy Many" paradigm for text-only familial models.
Ruyi2.5 extends this co-evolution discipline to the multimodal domain, introducing joint text-image understanding as a first-class capability across all familial members.
Concretely, Ruyi2.5 makes the following contributions. First, we propose a Shared-Backbone Multimodal Familial Model architecture in which a single training run yields multiple deployable sub-models of varying depth and width through parameter inheritance and flexible branching \citep{song2026ruyi2,chen2022knowledgeinheritance}. 
All familial members thus remain within the same semantic and capability space, enabling seamless handoff between edge and cloud inference stages. Empirical results confirm that this unified training scheme does not introduce systematic performance degradation relative to independently trained models of the same scale, while reducing total compute by avoiding redundant training runs.

Second, we instantiate the familial model architecture in Ruyi2.5-Camera, a structured two-stage pipeline for privacy-aware behavior recognition. At the edge, a lightweight familial member performs source-level de-identification by applying nonlinear mapping and stochastic noise injection to raw image frames, guided by information bottleneck theory \citep{shao2021learning,ye2024privacysemanticcommunications}. 
The resulting abstract feature vectors mathematically preclude reconstruction of the original visual content while preserving action-discriminative structure. 
On the cloud, a larger co-trained familial member performs deep behavior reasoning over these anonymized representations, handling occlusions, multi-person interactions, and long-horizon temporal dynamics that exceed the computational budget of the edge node.
This architecture realizes the transition from pixel-level video surveillance to de-identified behavior perception, enabling safety monitoring in the most privacy-sensitive environments.

Third, we propose Binary Prefix Policy Optimization (BPPO), a novel RLVR algorithm designed to accelerate multimodal long-chain reasoning. BPPO introduces structural constraints along two orthogonal axes: at the sample level, it retains only two representative responses per optimization group—one from the positive-reward stratum and one from the negative—eliminating redundancy while preserving the most discriminative learning signal; at the token level, it restricts gradient propagation to the response prefix, motivated by the observation that early generative steps in long-chain reasoning disproportionately determine the subsequent solution trajectory \citep{yu2025dapo}. The resulting objective function incorporates importance sampling ratios, group-relative advantage estimates, clipping stabilization, and KL regularization, maintaining full alignment with the RLVR paradigm while achieving a 2–3× training speedup over standard GRPO-based methods.

We evaluate Ruyi2.5 along two complementary axes. On standard multimodal benchmarks spanning visual question answering, fine-grained recognition, cross-modal reasoning, and task generalization, Ruyi2.5 performs on par with Qwen3-VL \citep{yang2025qwen3} models of equivalent scale, establishing competitive general-purpose multimodal capability. On scene-specific benchmarks for privacy-camera behavior recognition, Ruyi2.5-Camera substantially outperforms Qwen3-VL, confirming the practical value of familial model joint training, information-bottleneck anonymization, and edge-cloud collaboration under strict real-world constraints.

\section{Architecture}

\begin{figure}[]
    \centering
    \includegraphics[width=\textwidth]{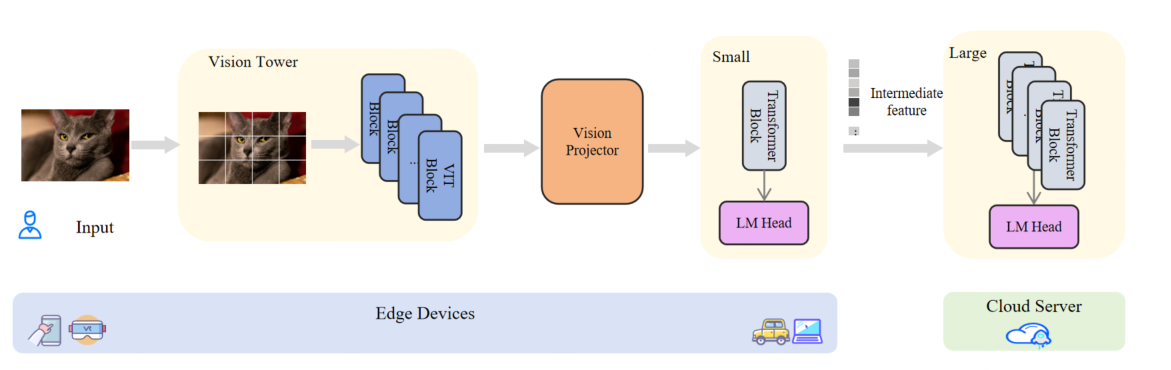}
    \caption{Schematic diagram of the Ruyi2.5 multimodal architecture}
    \label{ruyi_2_5}
\end{figure}

Ruyi2.5 is built upon a device-cloud collaborative multimodal architecture. As the latest evolution of the model family, Ruyi2.5 successfully transitions into a multimodal Vision-Language Model (VLM) based on the pure-text large language model Ruyi2 \citep{song2026ruyi2}, by introducing a vision encoder and a vision-language projection layer. Please refer to Figure \ref{ruyi_2_5} for the model architecture diagram. In terms of specific implementation, we conducted deep secondary development based on the open-source Qwen3-VL \citep{qwen3vl_techreport} architecture. Given an input image $\mathcal{I}$ and a text instruction $\mathcal{T}$, we first utilize a shared vision encoder $\mathcal{E}_{v}$ to extract visual representations, and align them to the language model's input space via a projection layer $\mathcal{P}_{v \to l}$. Subsequently, these are concatenated with the text features $\mathcal{E}_{t}(\mathcal{T})$ to form a unified cross-modal initial context $\mathbf{H}_0$:
$$\mathbf{H}_0 = \mathcal{P}_{v \to l}(\mathcal{E}_{v}(\mathcal{I})) \oplus \mathcal{E}_{t}(\mathcal{T})$$
The constructed $\mathbf{H}_0$ serves as the foundational sequence and is directly fed into the language model backend for deep semantic interaction and autoregressive reasoning. This design not only inherits the three-stage paradigm of mainstream VLMs but also seamlessly extends the device-cloud collaboration philosophy of the Ruyi series into the multimodal domain.

At the language model backend, Ruyi2.5 deeply integrates the theoretical essence of "AI Flow" \citep{an2025aiflowperspectivesscenarios}. Based on the three core principles within this theoretical framework—namely, the Law of Information Capacity\citep{yuan2026informationcapacityevaluatingefficiency}, the Law of Familial Model\citep{song2026theoreticalfoundationsscalinglaw}, and the Law of Multi-Model Collaboration\cite{lu2026lawmultimodelcollaborationscaling}—we innovatively constructed a device-cloud collaborative cascaded architecture of large and small models\citep{yuan2025task}. Relying on the unified semantic space guaranteed by the Law of Familial Model, this architecture can seamlessly adapt to system-level dynamic task routing mechanisms: a lightweight branch containing only a single Transformer block is deployed on the device side to handle basic visual perception and simple instructions with ultra-low latency; meanwhile, the full-scale large model branch is retained on the cloud side, responsible for the deep processing of complex logical tasks.

Overall, Ruyi2.5 achieves a unified closed loop at the architectural level, from feature extraction to device-cloud cascaded reasoning. Without the need to retransmit the original image, it truly realizes the free flow of computational power and the step-wise adaptation of cognitive depth between the device and cloud extremities. Please refer to Table \ref{tab:ruyi25_arch} for the specific configurations of network layers and hidden dimensions for Ruyi2.5 models of different scales.

\begin{table}
\centering
\renewcommand{\arraystretch}{1.2}
\caption{Model architecture details of Ruyi 2.5.}
\begin{tabular}{l c c c c}
\toprule
\multirow{2}{*}{\textbf{Model}} & \multicolumn{2}{c}{\textbf{Vision Tower}} & \multicolumn{2}{c}{\textbf{LM}} \\
\cmidrule(lr){2-3} \cmidrule(lr){4-5}
 & ViT Blocks & Hidden Size & Decoder Block(s) & Hidden Size \\
\midrule
Ruyi 2.5-1.9B & 27 & 1152 & 1  & 4096 \\
Ruyi 2.5-8B   & 27 & 1152 & 36 & 4096 \\
\bottomrule
\end{tabular}

\label{tab:ruyi25_arch}
\end{table}

\section{Post-training}

\subsection{Post-training Data}

To meet the core requirement of high-quality data for multimodal model training, we built a large-scale image resource pool with broad coverage and carefully refined annotations. Guided by the principle of \emph{strictly controlling quality boundaries while expanding semantic coverage}, we designed an efficient data-governance pipeline. We first aggregate visual assets from diverse sources at scale to establish sufficient breadth. We then apply multi-stage denoising and redundancy removal based on high-dimensional feature representations, ensuring that the resulting samples are highly curated and distributionally balanced. Beyond satisfying strict standards for visual quality, the dataset also supports a wide range of supervised fine-tuning (SFT) tasks, spanning basic visual question answering, complex STEM reasoning, and document understanding. This high-density supervision forms the core of our data-mixing strategy and effectively improves the model's generalization ability and perceptual robustness in open-domain scenarios.

\subsubsection{Construction of Multimodal SFT and RL Data Sources}

During the supervised fine-tuning (SFT) stage, the diversity and domain coverage of training data play a decisive role in shaping model capabilities. To this end, we systematically integrated and constructed a large-scale, multi-source multimodal instruction-tuning dataset. Its primary data format consists of image--text pairs, complemented by structured task annotations, thereby providing stable and information-rich cross-modal supervision signals.

In terms of task coverage, the dataset includes not only fundamental vision--language tasks such as visual question answering (VQA), image captioning, and fine-grained visual understanding, but also extends to task scenarios with greater cognitive complexity and deeper reasoning demands. Specifically, we introduce multimodal code generation tasks, enabling the model to generate or interpret program-related textual expressions conditioned on visual inputs. We also incorporate complex STEM problem-solving samples to strengthen the model's capabilities in structured knowledge utilization and logical reasoning. In addition, we place particular emphasis on cross-modal document understanding data, including reasoning over charts, technical documents, and academic PDFs with high information density, thereby enhancing the model's competence in realistic and complex document-centric scenarios.

At the level of domain distribution, the dataset expands the coverage of the visual-semantic space through cross-task and cross-domain organization, while improving the model's adaptability across a wide range of vision--language interaction settings. By introducing multi-level task structures and high knowledge density, the dataset provides more diverse and challenging training signals, which in turn improves the model's multimodal understanding, reasoning, and generation abilities in open-domain environments.

Furthermore, during the reinforcement learning (RL) stage, we construct a dedicated dataset for geometry-related mathematical problems, with a particular focus on plane geometry, solid geometry, and diagrammatic reasoning. This specialized data further enhances the model's ability in visually grounded mathematical derivation, spatial relation modeling, and multi-step logical problem solving.

\subsubsection{A Strict Data Filtering Pipeline}

To jointly optimize data quality and distributional breadth, we implement a strict staged data-governance and filtering strategy when constructing the large-scale corpus. First, at the basic quality-control stage, we define hard thresholds on low-level visual attributes such as spatial resolution, aspect ratio, and image integrity, and automatically remove low-quality, blurry, and corrupted samples. Next, at the stage of semantic distribution balancing, we extract global visual feature vectors for all samples and perform efficient hierarchical clustering in a high-dimensional embedding space. Based on the resulting cluster structure, we further conduct structured modeling of the overall data distribution.

To avoid the homogenization problem commonly induced by conventional random sampling, we do not adopt simple proportional sampling. Instead, we introduce a greedy sample-selection strategy based on distances in the feature space. Within each semantic cluster, candidate samples are selected by maximizing pairwise cosine distance, thereby improving the diversity and coverage of the final subset along semantic dimensions, entity types, and visual styles.

This strategy not only suppresses the over-concentration of high-frequency concepts, but also significantly increases the retention probability of long-tail concepts and rare scenarios during sampling. As a result, the final dataset exhibits a more balanced semantic distribution, providing more representative visual-semantic supervision signals for model training and improving the model's generalization ability and perceptual robustness in open-domain complex scenarios.

To further enhance the overall quality of supervisory signals during the SFT (Supervised Fine-Tuning) stage and mitigate interference from low-quality or biased data, we designed and implemented a two-stage data filtering mechanism targeting "instruction-response" pairs. On the instruction side, we employ a combination of heuristic rules and high-capability evaluation models to prune redundant, logically incoherent, or low-value queries that can be resolved through simple common sense, thereby ensuring that the retained samples possess high information density and reasoning complexity. On the response side, we focus on evaluating the accuracy and compliance of the outputs. By utilizing multi-dimensional assessment models, we filter out low-quality samples characterized by factual errors, hallucinatory tendencies, or linguistic disorganization. This process strictly selects responses with high instruction-following fidelity and rigorous articulation that align with human preferences, ensuring both the efficiency and stability of the supervised fine-tuning process at the data level.

\subsection{Post-training Strategy}

During post-training, we construct a highly representative full-scenario SFT dataset, aiming to push the model from basic visual perception toward higher-order cognitive reasoning. To preserve generality while maximizing the model's potential in complex domains, we adopt the following three key fine-tuning strategies.

\subsubsection{Cross-Domain Distribution Equalization}

To build a model with all-around multimodal understanding ability, our dataset takes captioning and general question answering as the central foundation, and is further expanded to cover optical character recognition (OCR), visual grounding, professional scientific knowledge, and advanced mathematical reasoning. During training, instead of naively mixing all data uniformly, we carefully control the proportion of each domain and adjust the relative weights of vertical domains throughout the fine-tuning stage. This equalization strategy effectively suppresses overfitting to individual high-frequency scenarios and mitigates potential domain bias, thereby ensuring robust generalization across diverse tasks.

\subsubsection{Staged Alignment of Vision and Language}

To deeply couple visual representation learning with multimodal logical reasoning, we design a staged evolutive training paradigm. The early stage focuses on spatial consistency alignment, using high-resolution image features and textual instructions to establish strong foundations in visual perception and attribute recognition. The training emphasis is then smoothly shifted toward a cognition-enhancement stage, where long-context reasoning data and cross-image relational tasks are introduced. This guides the model to evolve deeper semantic association and logical inference capabilities while preserving fine-grained perceptual sensitivity. Such a progressive alignment strategy effectively addresses the coordination challenge between the visual encoder and the large language model (LLM), allowing the final model to achieve a strong balance between fine-grained understanding and complex multimodal dialogue.

\subsubsection{Efficient Reinforcement Learning Training}

\emph{Binary Prefix Policy Optimization} (BPPO) is proposed to address two major inefficiencies in conventional group-based policy optimization: high intra-group sample redundancy and the high computational cost of long-sequence back-propagation. Results from our experimental analysis indicate that gradients induced by positive responses within the same group are highly similar, and the same redundancy also appears among negative responses, whereas the contrast between positive and negative responses provides the most informative optimization signal. BPPO therefore retains only one representative positive response and one representative negative response from each group for policy updates, reducing redundant gradient computation while preserving the essential learning signal.

In autoregressive reasoning, early response tokens play a dominant role in determining problem interpretation and the subsequent solution trajectory, whereas later tokens often extend an already established reasoning path. Accordingly, gradient propagation is restricted to the response prefix rather than the full generated sequence. By combining binary representative response selection with prefix-constrained optimization, BPPO significantly reduces reinforcement learning cost while maintaining the core training benefits of GRPO. Based on this design, the BPPO objective can be formalized as:
\begin{equation}
J_{\mathrm{BPPO}}(\theta)=
\mathbb{E}\Bigg[
\frac{1}{2}\sum_{i\in \mathcal{S}(q)}
\frac{1}{\sum_{t=1}^{|o_i|} m_{i,t}^{(n)}}
\sum_{t=1}^{|o_i|}
m_{i,t}^{(n)}
\min\!\left(
\rho_{i,t}\hat{A}_{i,t},
\operatorname{clip}(\rho_{i,t},1-\varepsilon,1+\varepsilon)\hat{A}_{i,t}
\right)
-\beta D_{\mathrm{KL}}\!\left(\pi_{\theta}\,\|\,\pi_{\mathrm{ref}}\right)
\Bigg],
\end{equation}
where $\mathcal{S}(q)$ denotes the binary response set selected for query $q$, containing one positive response and one negative response; $o_i$ denotes the $i$-th response sequence; $m_{i,t}^{(n)}$ is the prefix mask that restricts gradient updates to the first $n$ tokens of response $o_i$; $\rho_{i,t}$ denotes the importance sampling ratio at token $t$ of response $o_i$; $\hat{A}_{i,t}$ is the advantage term derived from group-relative rewards; $\varepsilon$ is the clipping coefficient; and $\beta$ is the weight of the KL regularization term.

\subsection{Post-training Evaluation}

We evaluate the Ruyi2.5 familial models on behavior recognition tasks in privacy-sensitive and complex scenarios, focusing on their practical performance in real-world applications. To this end, we trained a dedicated \textbf{Ruyi2.5-Camera} model based on the Ruyi2.5 family to handle complex behavior recognition under camera surveillance.

As shown in Table 2, Ruyi2.5-Camera demonstrates robust and solid multimodal reasoning capabilities on the targeted behavior recognition benchmarks. As the main model in the family, it significantly outperforms the mainstream Qwen3-VL model on key scene-level metrics, illustrating its ability to understand and discriminate complex behaviors in privacy-constrained environments. Experimental results indicate that Ruyi2.5-Camera can reliably handle specific behavior recognition tasks while demonstrating the technical strength and practical applicability of the Ruyi2.5 familial models in real deployment scenarios.

\begin{table}[htbp]
\centering
\caption{Comparison of behavior recognition performance in privacy-sensitive and complex surveillance scenarios. Ruyi2.5-Camera-8B achieves the best results across all evaluation metrics, demonstrating the effectiveness of the Ruyi2.5 family for practical real-world deployment.}
\label{tab:behavior_recognition_results}
\begin{tabular}{lcccc}
\hline
Model & Accuracy & Precision & Recall & F1-Score \\
\hline
Qwen3-VL-8B         & 81.67 & 82.93 & 84.10 & 82.77 \\
Ruyi2.5-Camera-1.9B & 77.67 & 84.33 & 70.30 & 76.47 \\
Ruyi2.5-Camera-8B   & \textcolor{red}{93.33} & \textcolor{red}{95.50} & \textcolor{red}{91.87} & \textcolor{red}{93.67} \\
\hline
\end{tabular}
\end{table}


\section{Conclusion }

We presented Ruyi2.5, a multimodal familial model that resolves the long-standing tension between privacy protection and semantic behavior understanding in surveillance scenarios. Building on the AI Flow framework, Ruyi2.5 extends the Ruyi2 shared-backbone paradigm to the multimodal domain, enabling models of different scales to co-evolve within a single training pipeline and operate seamlessly across device, edge, and cloud tiers.

The Ruyi2.5-Camera pipeline decomposes recognition into two stages: irreversible source-level de-identification at the edge, grounded in information bottleneck theory, and deep multimodal behavior reasoning on the cloud. This design provides formal privacy guarantees—precluding image reconstruction—while preserving the semantic richness needed to detect falls, altercations, and other risk behaviors in highly constrained environments. Experiments confirm that Ruyi2.5 matches Qwen3-VL on standard multimodal benchmarks, and Ruyi2.5-Camera substantially surpasses it under privacy-constrained real-world conditions.

Binary Prefix Policy Optimization (BPPO) further contributes a training-efficiency advance: by selecting only binary representative responses per group and concentrating gradient updates on response prefixes, BPPO achieves a 2–3× speedup over standard GRPO without degrading final policy quality.

Looking ahead, the Ruyi series will advance along four directions: deeper shared foundations for richer modality mixtures; finer capability stratification across tiers; adaptive edge-cloud routing under dynamic network conditions; and closed-loop self-improvement from deployment feedback. These directions collectively point toward scalable, privacy-respecting intelligent perception systems suitable for healthcare, public safety, and beyond.

\bibliographystyle{plainnat}
\bibliography{paper}

\end{document}